\documentclass[conference,compsoc]{IEEEtran}
\ifCLASSINFOpdf
\else
   \usepackage[dvips]{graphicx}
\fi
\usepackage{url}
\usepackage{graphicx}
\usepackage{subcaption}
\usepackage{amsmath}
\usepackage{amssymb}
\usepackage{booktabs}
\begin{document}

\title{Why Commodity WiFi Sensors Fail at Multi-Person Gait Identification: A Systematic Analysis Using ESP32}

\author{\IEEEauthorblockN{Oliver Custance, Saad Khan, Simon Parkinson}
\IEEEauthorblockA{Department of Computer Science\\
University of Huddersfield\\
UK\\
Email: oliver.custance@hud.ac.uk, saad.khan@hud.ac.uk, and s.parkinson@hud.ac.uk}
}


\markboth{IEEE S\&P Conference Submission}{}
\maketitle

\begin{abstract}
WiFi Channel State Information (CSI) has shown promise for single-person gait identification, raising interest in its use for contactless biometrics, continuous authentication, and passive identification. However, the feasibility of multi-person identification on low-cost commodity devices remains unclear. A critical question is whether weak multi-person performance is primarily an algorithmic limitation, or whether it reflects a more fundamental sensing ceiling on commodity WiFi hardware. We address this question through a systematic empirical study using commodity ESP32 WiFi sensors. We evaluated six different signal separation methods—FastICA, SOBI, PCA–ICA, NMF, Wavelet, and Tensor decomposition—across seven scenarios spanning 1–10 people in both controlled and realistic indoor environments. To investigate beyond classification accuracy, we introduce three diagnostic metrics: intra-subject variability (ISV), inter-subject distinguishability (ISD), and performance degradation rate (PDR). In all methods, performance remains moderate (39\%-56\% accuracy), with limited evidence that algorithmic choice alone solves the problem. The best-performing method, NMF, reaches 56\% accuracy, while all methods exhibit extremely high feature-space overlap (97–99\%), unstable within-subject representations, and marked environmental sensitivity. These findings suggest that, under commodity ESP32 CSI constraints, dense multi-person gait identification is limited more by sensing quality and spatial diversity than by the chosen separation algorithm.  Our results have direct implications for security and privacy: they call into question the practicality of commodity WiFi CSI as a robust multi-user biometric primitive for authentication, while also placing important bounds on the passive identification capabilities achievable with low-cost off-the-shelf WiFi hardware.

\end{abstract}


\IEEEpeerreviewmaketitle

\section{Introduction}
WiFi-based human sensing has emerged as a promising technology for non-intrusive activity recognition and identification~\cite{ma2019wifi, liu2019wireless}. Unlike camera-based systems that raise privacy concerns, or wearable sensors that require user cooperation, WiFi sensing leverages existing wireless infrastructure to infer human behaviour through Channel State Information (CSI)~\cite{wang2015understanding}. CSI captures fine-grained information from the physical-layer on signal propagation, making it sensitive to environmental changes caused by human movement~\cite{yousefi2017survey}. Beyond activity recognition, WiFi gait sensing has been increasingly discussed as a candidate biometric modality for security applications such as continuous authentication and identity-aware smart environments. At the same time, the same sensing capability raises privacy concerns because passive RF sensing could, in principle, enable covert identification without cameras or active user participation. For the security and privacy community, the key question is therefore not only whether WiFi-based gait identification can work in idealised laboratory conditions but whether commodity, low-cost devices can support such identification reliably in realistic multi-person environments. If they can, commodity WiFi may represent a practical authentication primitive and a significant privacy threat. If they cannot, both deployment claims and threat models need to be reconsidered.

Single-person gait identification using WiFi CSI has demonstrated strong performance in controlled settings, with multiple studies reporting high identification accuracy~\cite{zhang2018crosssense, wang2018csi, ding2020wihi, custance2023classifying,custance2026parameter,11298190}. These systems exploit biomechanical regularities in walking that modulate CSI amplitude and phase as people move through the wireless field~\cite{jiang2018towards}. This success motivates the extension to multi-person scenarios, where multiple individuals are present simultaneously~\cite{xiao2012fifs, zhang2021widar3}. However, multi-person gait identification introduces a fundamental challenge: \emph{signal separation}. When multiple people move simultaneously, their gait signatures interfere and mix in the received CSI, creating a superposition that must be decomposed before individual identification can occur~\cite{shi2017smart}.

Blind source separation approaches such as Independent Component Analysis (ICA), tensor decomposition, and wavelet transforms provide potential algorithmic tools~\cite{abuhoureyah2024multi, tokcan2025tensor, hernandez2022wifi}, but multi-person WiFi gait identification remains relatively underexplored. The limited existing work suggests a concerning pattern: systems either depend on complex sensing configurations, such as modified firmware and specialised antenna arrays~\cite{zhang2021gaitsense}, or they achieve only modest performance when using commodity hardware~\cite{li2024spacebeat, rao2024novel}. This raises a key unresolved question: \emph{is poor multi-person performance primarily an algorithmic limitation, or does it stem from fundamental constraints of commodity WiFi devices?}

Understanding where this bottleneck lies has direct implications for the direction of research and deployment. If algorithmic improvements are sufficient, the effort should focus on improved separation and identification models. If hardware limitations dominate, then incremental algorithmic refinement may have diminishing returns, and reliable multi-person identification may require richer sensing modalities such as massive MIMO or mmWave systems~\cite{wang2024rtmp, ganesan2021clustering, ni2022gait, xia2021person, zeng2022multi}. In practical terms, the question is whether multi-person gait identification is a tractable extension of single-person CSI sensing on low-cost embedded devices, or whether the sensing problem itself is underdetermined.

\begin{figure*}[ht]
    \centering
    \includegraphics[width=\linewidth]{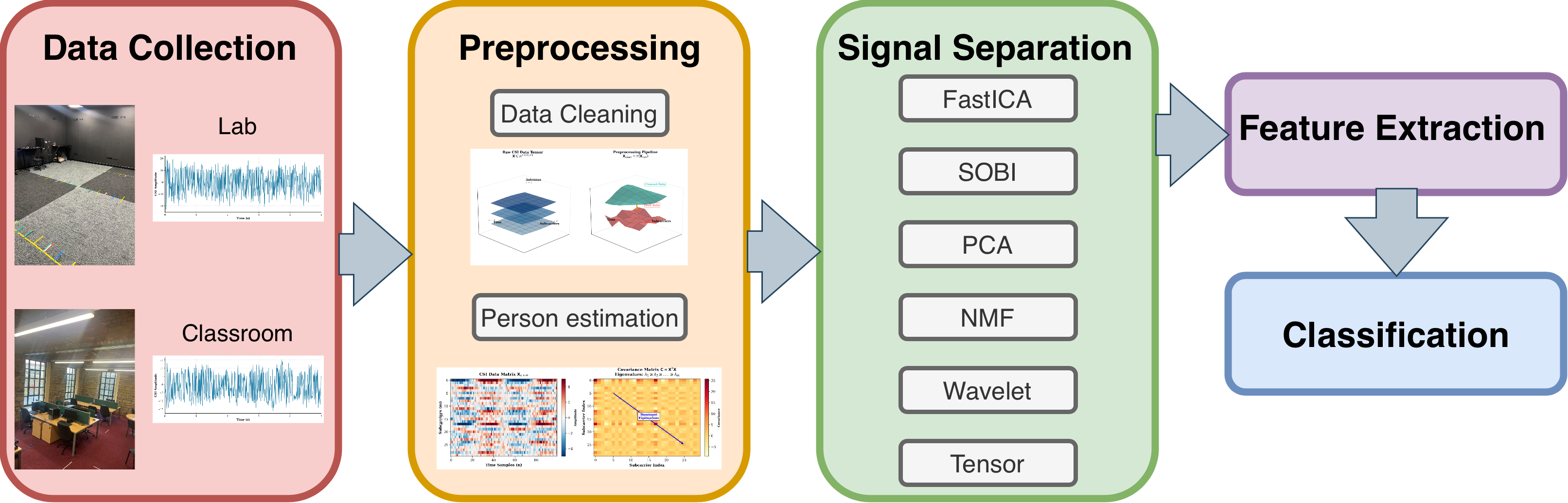}
    \caption{Overview of the workflow, from ESP32 CSI capture through preprocessing and person-count estimation, to separation of mixed observations and final identification using feature-based classification}
    \label{fig:pipeline}
\end{figure*}

To address this gap, we conducted a systematic empirical study using commodity ESP32 WiFi sensors. Unlike previous systems that depend on expensive Intel 5300 NICs with modified drivers or more elaborate configurations, ESP32 represents a simple, low-cost, and deployable platform. This makes the study intentionally demanding: the sensing stack is constrained, and any gains must emerge from the data and algorithms rather than from additional infrastructure. We therefore compare six diverse separation methods under the same acquisition conditions and evaluate them using both standard classification metrics and diagnostic measures designed to explain \emph{why} methods fail or succeed. The overall experimental and analysis workflow is shown in Fig.~\ref{fig:pipeline}. The contributions of this paper are as follows:
\begin{itemize}
    \item We provide a systematic comparison of six blind source separation strategies--FastICA, SOBI, PCA–ICA, NMF, Wavelet, and Tensor decomposition for multi-person gait identification using commodity ESP32 CSI across seven scenarios spanning 1–10 people.
    
    \item  We introduce three diagnostic metrics. These are: intra-subject variability (ISV), inter-subject distinguishability (ISD), and performance degradation rate (PDR) to characterise representation quality and failure modes beyond headline accuracy.

    \item We show that all evaluated methods remain in a high-overlap, weak-separability regime under commodity ESP32 sensing constraints, suggesting that sensing limitations dominate algorithmic differences in dense multi-person setting

    \item We discuss the security and privacy implications of this result, showing that commodity WiFi CSI is currently a weak foundation for robust multi-user gait biometrics in authentication settings, while also limiting the passive identification capability that low-cost commodity devices can realistically provide in crowded environments.
\end{itemize}

\section{Related Work}
This section reviews prior work most relevant to this study. Since the aim here is not to propose a new identification architecture but to establish practical limits on commodity sensing, the discussion is structured around: (i) multi-person WiFi sensing, (ii) blind source separation methods for mixed observations, and (iii) constraints that limit separability in practice.

Multi-person sensing with WiFi CSI is fundamentally harder than single-person identification because measurements contain mixtures of multiple simultaneous motions, and downstream inference depends on whether user-specific contributions can be separated reliably. Across the literature, strong multi-user results typically come from increasing observation diversity, for example through multi-antenna reception, multiple links, or wider effective bandwidth, and pairing it with separation or decomposition models that match the signal structure. For example, MultiSense frames multi-person WiFi respiration monitoring as a blind source separation problem and reports accurate monitoring for up to four persons~\cite{zeng2020multisense}, while WiMUSE reports substantial reductions in respiration-rate error relative to FFT/MUSIC-style baselines~\cite{yi2025multi}. Beyond physiology, MultiTrack reports accurate localisation and activity recognition by exploiting multiple links and 5\,GHz channel splicing to separate per-user reflections~\cite{tan2021multi}.

For gait identification specifically, WiWalk targets the dual-user case using a learning-based time--frequency separation approach and reports high identification accuracy for small user groups, decreasing as the number of users increases~\cite{ou2022wiwalk}. Most directly, MUID reports high average accuracies for 2-person and 3-person gait identification across three environments~\cite{wang2025muid}, but also shows strong sensitivity to separability conditions such as narrow user spacing and relies on increased spatial diversity via array-style reception and angle-of-arrival driven separation. Together, these findings motivate our position: rather than assuming separability and then refining classifiers, the first question is whether a highly constrained commodity CSI platform can support stable multi-person separation in the first place.

A recent study presents a complete WiFi-based gait identification pipeline, incorporating CSI denoising and subcarrier selection, gait segmentation via moving-variance analysis, walking-direction estimation using Channel Power Distribution (CPD), and identity classification through a hybrid Transformer architecture~\cite{11298190}. The system combines convolutional layers, residual LSTM blocks, and multi-head self-attention, along with a feature fusion strategy integrating raw subcarrier amplitudes, time–frequency descriptors, and autoencoder-derived latent features. Evaluated on a 25-subject ESP32 dataset (WiStride) and two public datasets (MultiEnvironment and HWDD), the approach achieves high identification accuracy (up to 97.9\%) and demonstrates robustness across environments and participant scales, outperforming standard baselines such as CNNs, LSTMs, and SVMs. The work further extends to open-set verification using learnt embeddings, reporting strong biometric performance (AUC 96.2\%, EER 10.8\%).

Blind source separation (BSS) tackles the common situation where the measured signal is not a single clean source, but a superposition of several underlying sources, and the way they are mixed is not known in advance. In multi-user WiFi identification, this helps explain why single-person methods often break down: when two or more people move at the same time, the CSI contains a mixture of their effects. This mixing can blur or hide identity-specific patterns unless the system can explicitly separate users, or makes stronger assumptions that make the mixture easier to disentangle~\cite{wei2025survey}.

Recent WiFi and communications studies also show that even when separation is explicitly attempted, performance is strongly bounded by measurement diversity and real hardware effects. In recent work, the authors apply an ICA-family approach (JADE) to separate mixed activities of two people from the CSI amplitude and report average recognition accuracies in the mid-70\% range between volunteer groups, with separation becoming poor under unfavourable geometry~\cite{li2020valid}. From a signal-processing perspective, in one work, the authors demonstrate that ICA-based separation can outperform least-squares baselines in low-SNR regimes, while also highlighting practical constraints such as sensitivity to nonlinearity and the need for sufficient sample support as mixtures become more complex~\cite{fouda2021blind}. These findings are why we keep the sensing setup fixed and test separation methods within the same commodity constraints: the goal is to determine whether failures come from the separation algorithm itself, or simply from the CSI not containing enough usable information to separate people in the first place.

A recurring limitation in WiFi gait literature is that reported gains often depend on sensing assumptions that are difficult to satisfy in deployment. Systems that rely on calibrated phase, multiple synchronised links, or array-style angle estimation can improve separation, but these requirements are often unavailable on low-cost embedded hardware. On commodity devices, the number of antennas, the available subcarriers, and the stability of the reported CSI all constrain what can be separated in practice. This makes it unlikely that multi-person gait identification on ESP32-class hardware can simply inherit the gains reported on richer platforms. Our study therefore treats multi-person separability as an empirical question, not an assumption.

Most prior work evaluates WiFi gait sensing either as an activity-recognition problem or as a biometric identification problem under controlled sensing assumptions. However, for a security and privacy audience, the central issue is practical feasibility: whether low-cost commodity WiFi hardware is sufficient to support reliable identification in settings relevant to authentication or passive surveillance. Our work differs from prior studies by treating separability under commodity constraints as the primary question, rather than assuming that accurate identification is achievable and focusing only on classifier improvements.

\section{Methods}
This section describes the experimental methodology, including the ESP32 sensing configuration, the experimental conditions used to elicit multi-person gait mixtures, and the analysis pipeline used to evaluate separability under commodity constraints. We intentionally restrict the analysis to amplitude-based commodity ESP32 CSI, rather than phase-stable CSI obtainable from specialised NIC firmware, because our goal is to characterise what is achievable under realistic low-cost deployment constraints. This choice should therefore be interpreted as an evaluation of commodity practicality rather than an upper bound on WiFi sensing in general.

\subsection{ESP32 Hardware and CSI Configuration}
WiFi channel state information (CSI) is collected using commodity ESP32 devices operating in the 5~GHz band under IEEE 802.11n. For each packet, each ESP32 reports CSI amplitude measurements from $k=3$ antennas over $m=52$ information-bearing orthogonal frequency-division multiplexing (OFDM) subcarriers, obtained after removing the null and guard subcarriers from the original 64-subcarrier system~\cite{Hern2006:Lightweight}. 

For a single trial, the resulting CSI amplitude sequence is represented as a third-order tensor
\begin{equation}
\mathbf{X} \in \mathbb{R}^{n \times m \times k},
\end{equation}
where $n$ is the number of time samples in the recording window, $m=52$ is the number of retained information-bearing subcarriers, and $k=3$ is the number of antennas. Equivalently, each element $x_{t,s,a}$ of $\mathbf{X}$ denotes the CSI amplitude measured at time index $t \in \{1,\dots,n\}$, subcarrier index $s \in \{1,\dots,m\}$, and antenna index $a \in \{1,\dots,k\}$.

The design choice to use ESP32 CSI amplitude, rather than higher-resolution phase-stable CSI obtained from specialised NIC firmware or software-defined radio front-ends, is intentional. It reflects the sensing constraints of widely available commodity devices and therefore directly probes the practical information content available for multi-person separation under such constraints.

\subsection{Experimental Conditions and Scenarios}
Data collection was conducted in two environments: a controlled Laboratory (7 \, m $\times$ 7 \, m), designed to reduce uncontrolled environmental variation, and a realistic Classroom (5.4 \, m $\times$ 9 \, m), containing furniture and ambient WiFi traffic to better reflect multipath conditions similar to deployment. Within each environment, ESP32 transceivers were positioned to maximise spatial diversity across three spatial streams. This resulted in TX--RX separation distances of 5.8--7.3\,m in the Laboratory and 1.4--9.0\,m in the Classroom.

Ten participants (ages 20--40) contributed walking trials across seven experimental scenarios (A--G), spanning person counts from 1--10. These scenarios include single-person conditions (A, D), two-person conditions (B, E), five-person conditions (C, F), and a ten-person configuration (G). Throughout the sessions, participants wore different clothing but did not carry electronic devices.

\begin{figure}[ht]
    \centering
    \begin{subfigure}[t]{0.49\linewidth}
        \centering
        \includegraphics[width=\linewidth]{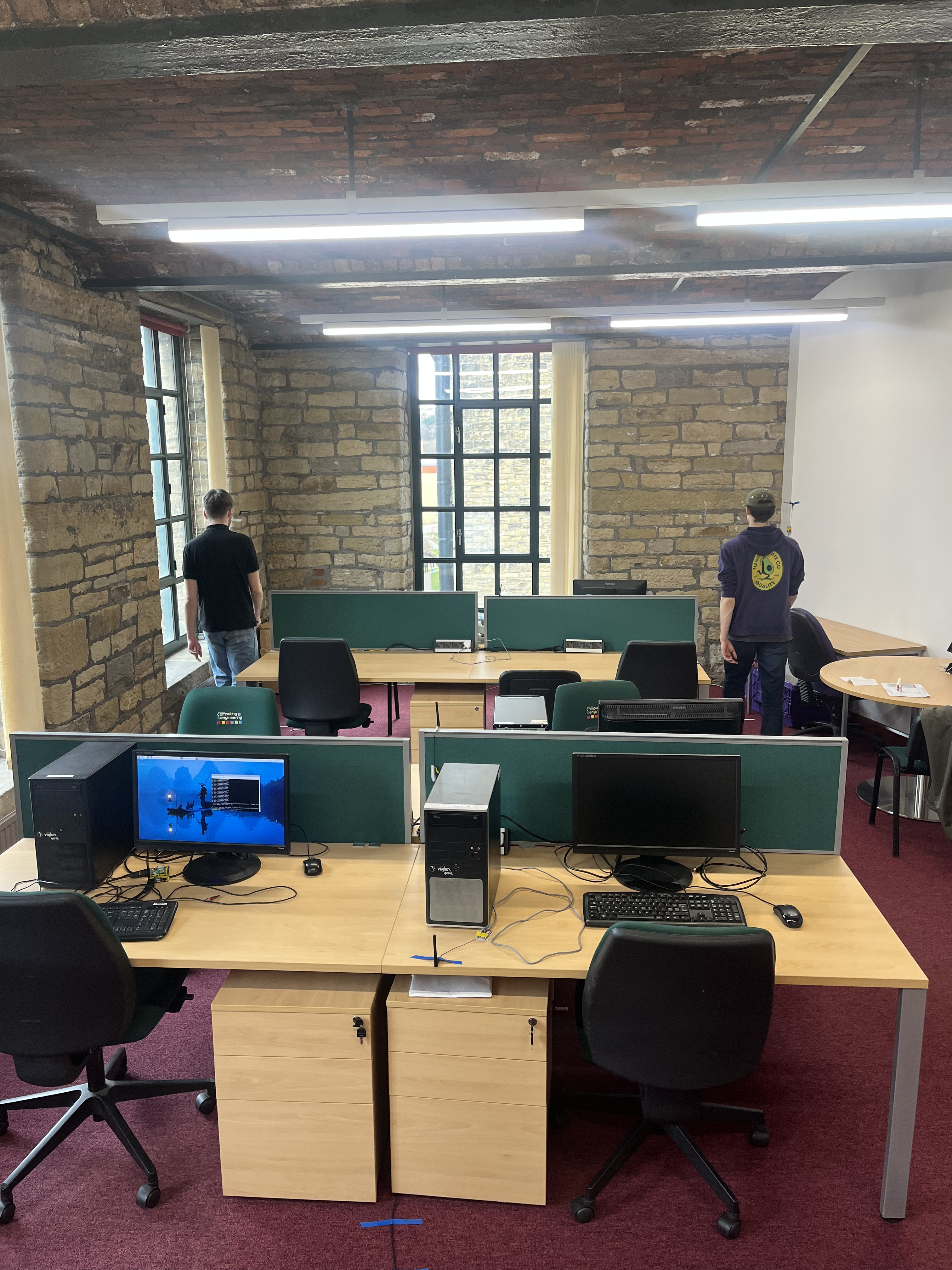}
        \caption{Lab environment with taped floor markers used to control starting positions and spacing.}
        \label{fig:study4_lab}
    \end{subfigure}
    \hfill
    \begin{subfigure}[t]{0.49\linewidth}
        \centering
        \includegraphics[width=\linewidth]{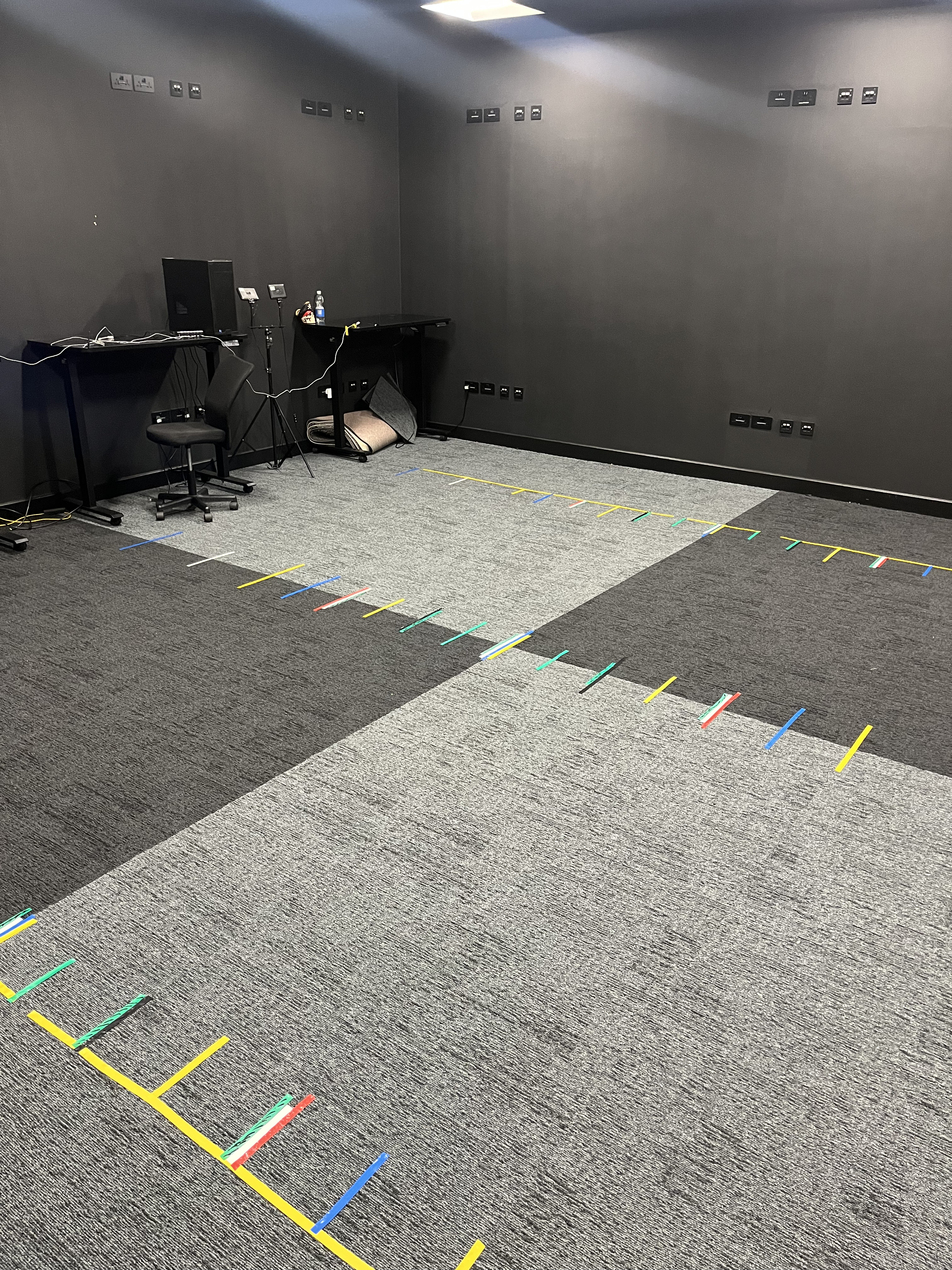}
        \caption{Classroom environment with marked walking paths used to standardise trajectories across trials.}
        \label{fig:study4_classroom}
    \end{subfigure}
    \caption{Study~4 data collection environments and floor markings used to standardise participant spacing (Lab) and walking trajectories (Classroom).}
    \label{fig:study4_environments}
\end{figure}

To keep the multi-person conditions repeatable, floor markings were used to standardise where participants started and how they moved through the sensing area. In the Laboratory, coloured tape marks indicate fixed standing positions and spacing references, allowing inter-person distances to be kept consistent and measured across trials. In the Classroom, tape markings define multiple walking paths and turning points, so repeated trials follow comparable trajectories despite the more cluttered environment. The two data collection environments and their corresponding floor markings are shown in Fig.~\ref{fig:study4_environments}.

These experimental conditions are intended to surface key multi-person failure modes relevant to gait identification on commodity CSI:
\begin{itemize}
    \item \textbf{Person count:} increasing the number of subjects walking simultaneously to increase the complexity of the mixture.
    \item \textbf{Trajectory variation:} repeated walking trials along predefined paths to yield comparable mixtures across repetitions while still allowing natural gait variation.
    \item \textbf{Inter-person spacing and occlusion:} scenarios include conditions where subjects can become close and partially occluded, increasing the likelihood of overlapping micro-Doppler and multipath signatures.
\end{itemize}

For clarity and reproducibility, Table~\ref{tab:study4_scenarios} summarises the scenario structure.

\begin{table*}[t]
\centering
\caption{Experimental scenarios and trial counts.}
\label{tab:study4_scenarios}
\small
\begin{tabular*}{\textwidth}{@{\extracolsep{\fill}} l l c c p{0.45\textwidth} @{}}
\toprule
\textbf{Scenario} & \textbf{Environment} & \textbf{\# People} & \textbf{Trials} & \textbf{Condition} \\
\midrule
A & Lab       & 1  & 30 & Single-person walking \\
B & Lab       & 2  & 80 & Two-person walking \\
C & Lab       & 5  & 80 & Five-person walking \\
D & Classroom & 1  & 30 & Single-person walking \\
E & Classroom & 2  & 80 & Two-person walking \\
F & Classroom & 5  & 80 & Five-person walking \\
G & Classroom & 10 & 6  & One person walking at a time; others seated \\
\bottomrule
\end{tabular*}
\end{table*}

\subsection{Data Preprocessing}
Each trial is first reduced to the 52 information-bearing subcarriers so that subsequent processing is performed on a consistent CSI representation. Per-subcarrier $z$-score normalisation is then applied to reduce sensitivity to absolute amplitude scale and to stabilise comparisons across subcarriers and trials. Specifically,
\begin{equation}
\mathbf{X}_{\text{norm}}(i,j,a) = \frac{\mathbf{X}(i,j,a)-\mu_j}{\sigma_j},
\end{equation}
where $\mathbf{X}_{\text{norm}} \in \mathbb{R}^{n \times m \times K}$ is the normalised CSI tensor, $i \in \{1,\dots,n\}$ indexes time samples, $j \in \{1,\dots,m\}$ indexes subcarriers, and $a \in \{1,\dots,K\}$ indexes antennas, with $m=52$ and $K=3$. Here, $\mu_j$ and $\sigma_j$ are the mean and standard deviation of subcarrier $j$ within the trial, computed across all time samples and antennas.

Finally, samples are aligned in time using nearest-neighbour timestamp matching. Where short gaps are present (less than 2\% of samples), forward--backward filling is applied to ensure a continuous time series for downstream person-count estimation and separation.

\subsection{Person Count Estimation}
The number of active sources is estimated, i.e., the effective number of contributors present in the CSI mixture, using an eigenvalue-based source enumeration method~\cite{tian2024source}. After reshaping the CSI tensor into a matrix $\widetilde{\mathbf{X}} \in \mathbb{R}^{n \times 52}$ so that each time sample is represented by a 52-dimensional subcarrier feature vector, the sample covariance matrix is computed as
\begin{equation}
\mathbf{C} = \frac{1}{n}\widetilde{\mathbf{X}}^T\widetilde{\mathbf{X}} \in \mathbb{R}^{52 \times 52},
\end{equation}
where $n$ is the number of time samples in the trial and $\mathbf{C}$ is the subcarrier covariance matrix. The eigendecomposition is then performed as
\begin{equation}
\mathbf{C} = \mathbf{U}\mathbf{\Lambda}\mathbf{U}^T,
\end{equation}
where $\mathbf{U}$ is the eigenvector matrix and $\mathbf{\Lambda}=\mathrm{diag}(\lambda_1,\lambda_2,\dots,\lambda_{52})$ is the diagonal matrix of ordered eigenvalues satisfying $\lambda_1 \geq \lambda_2 \geq \cdots \geq \lambda_{52}$. Intuitively, larger eigenvalues correspond to directions that explain more of the variance in the observed mixtures; as the number of active contributors increases, more eigen-directions are required to capture the dominant structure in the data.

Following~\cite{tian2024source}, the estimated source count $\hat{p}$ is set to the smallest integer $r$ such that the cumulative explained variance exceeds 95\%:
\begin{equation}
\hat{p} = \min \left\{ r \in \{1,\dots,52\} : \frac{\sum_{i=1}^{r}\lambda_i}{\sum_{i=1}^{52}\lambda_i} \geq 0.95 \right\}.
\end{equation}

\subsection{Signal Separation Methods}
Six blind source separation families are evaluated under the same sensing inputs to separate algorithmic sensitivity from sensing limitations. These six methods were selected to span distinct assumptions about mixture structure—statistical independence, second-order temporal structure, low-rank conditioning, non-negativity, multi-resolution decomposition, and tensor factorisation, thereby allowing us to test whether the observed limitations reflect method-specific weaknesses or a broader sensing bottleneck.

\subsubsection{FastICA}
FastICA assumes that the observed CSI features are linear mixtures of statistically independent non-Gaussian sources~\cite{liu2022packaged}. In the implementation, each trial is represented as a data matrix $\mathbf{X}\in\mathbb{R}^{n\times d}$, where $n$ is the number of time samples in the trial and $d$ is the number of CSI features per sample. Before separation, features are standardised to zero mean and unit variance to reduce scale effects across subcarriers. The PCA whitening is then applied and the data is reduced to dimensions $\hat{p}$, where $\hat{p}$ denotes the estimated number of active sources.

FastICA then maximises the negentropy-inspired contrast function
\begin{equation}
J(\mathbf{w}) = \left[E\{G(\mathbf{w}^{T}\mathbf{x})\} - E\{G(\nu)\}\right]^2,
\end{equation}
where $\mathbf{w}$ is a projection vector, $\mathbf{x}$ is a whitened observation vector, $E\{\cdot\}$ denotes expectation, $\nu\sim\mathcal{N}(0,1)$ is a standard Gaussian random variable, and $G(u)=(1/a)\log\cosh(au)$ with $a>0$ a constant controlling the nonlinearity. The separated components are then obtained as
\begin{equation}
\mathbf{S}_{\text{ICA}} = \mathbf{W}\tilde{\mathbf{X}},
\end{equation}
where $\tilde{\mathbf{X}}$ is the whitened data matrix and $\mathbf{W}$ is the ICA demixing matrix. Here, $\mathbf{S}_{\text{ICA}}$ denotes the matrix of recovered independent components.

\subsubsection{SOBI}
Second-Order Blind Identification (SOBI) is a blind source separation method that takes advantage of temporal structure by diagonalizing a set of time-delayed covariance matrices jointly~\cite{liu2022remove}. Let $\mathbf{x}(t)\in\mathbb{R}^{p}$ denote the multichannel observation vector at time index $t$, where $p$ is the number of retained features or channels. SOBI forms the lagged covariance matrices
\begin{equation}
\mathbf{R}_{\tau} = \mathbb{E}\{\mathbf{x}(t)\mathbf{x}(t-\tau)^T\}, \quad \tau \in \{\tau_1,\ldots,\tau_L\},
\end{equation}
where $\mathbf{R}_{\tau}\in\mathbb{R}^{p\times p}$ is the covariance matrix in lag $\tau$, $\mathbb{E}\{\cdot\}$ denotes the expectation, and $\tau_1,\ldots,\tau_L$ are the time delays selected in the samples. After centring and whitening, the unmixing matrix $\mathbf{W}$ is estimated by minimising the total off-diagonal energy across lags using iterative joint diagonalisation. In this work, $L=10$ lags are used, uniformly distributed between 1 and 50 samples.

\subsubsection{PCA--ICA Hybrid Separation}
In the PCA--ICA pipeline, mixed CSI is separated using a two-stage approach: PCA is first used to condition the data by reducing redundancy and preserving dominant variance directions, and ICA is then applied to attempt source separation~\cite{ji2021qualitative}. Starting from the normalised CSI matrix $\mathbf{X}_{\text{norm}}\in\mathbb{R}^{n\times d}$, where $n$ is the number of time samples in the trial and $d$ is the number of CSI features per sample, the sample covariance matrix is calculated as
\begin{equation}
\mathbf{C}=\frac{1}{n-1}\mathbf{X}_{\text{norm}}^{T}\mathbf{X}_{\text{norm}},
\end{equation}
where $\mathbf{C}\in\mathbb{R}^{d\times d}$ is the covariance matrix of the normalised features. Its eigendecomposition is
\begin{equation}
\mathbf{C}=\sum_{i=1}^{d}\lambda_i\mathbf{u}_i\mathbf{u}_i^{T},
\end{equation}
where $\lambda_i$ is the $i$th eigenvalue and $\mathbf{u}_i\in\mathbb{R}^{d}$ is the corresponding eigenvector. The trial is then projected onto the leading $\hat{p}$ eigen-directions:
\begin{equation}
\mathbf{Z}=\mathbf{X}_{\text{norm}}\mathbf{U}_{\hat{p}},
\end{equation}
where $\hat{p}$ is the number of retained components and $\mathbf{U}_{\hat{p}}=[\mathbf{u}_1,\ldots,\mathbf{u}_{\hat{p}}]\in\mathbb{R}^{d\times \hat{p}}$. Here, $\mathbf{Z}\in\mathbb{R}^{n\times \hat{p}}$ denotes the PCA-reduced representation. FastICA is then applied to $\mathbf{Z}$ to recover statistically independent components. In practice, PCA acts as a conditioning step by decorrelating the data and filtering variance in a lower-dimensional space, while ICA is responsible for the actual unmixing attempt.

\subsubsection{NMF}
Non-negative Matrix Factorisation (NMF) decomposes non-negative CSI data into an additive parts-based representation~\cite{torabi2023new}. Let $\mathbf{X}^{+}\in\mathbb{R}_{\geq 0}^{n\times d}$ denote the non-negative CSI data matrix, where $n$ is the number of time samples and $d$ is the number of CSI features per sample. NMF approximates $\mathbf{X}^{+}$ as the product of two non-negative matrices, $\mathbf{W}\in\mathbb{R}_{\geq 0}^{n\times r}$ and $\mathbf{H}\in\mathbb{R}_{\geq 0}^{r\times d}$, by solving
\begin{equation}
\min_{\mathbf{W},\mathbf{H}} \|\mathbf{X}^{+}-\mathbf{W}\mathbf{H}\|_{F}^{2} + \alpha(\|\mathbf{W}\|_{1}+\|\mathbf{H}\|_{1}),
\end{equation}
where $r$ is the factorisation rank, $\|\cdot\|_{F}$ denotes the Frobenius norm, $\|\cdot\|_{1}$ denotes the elementwise $\ell_{1}$-norm, and $\alpha \geq 0$ is a sparsity regularisation parameter. Here, $\mathbf{W}$ contains temporal activations and $\mathbf{H}$ contains basis vectors. The non-negativity constraint is a natural fit for amplitude-only CSI because it avoids cancellation between signed components and encourages an additive decomposition of mixed observations.

\subsubsection{Wavelet Transform}
The Wavelet pipeline uses a multi-resolution decomposition of each CSI time series using a level-$L$ discrete wavelet transform with Daubechies-4 wavelets, where $L=4$ is the maximum decomposition level~\cite{osadchiy2021signal}. Let $x_j(t)$ denote the $j$th CSI feature time series, where $j$ indexes the feature channel. Each time series is decomposed as
\begin{equation}
x_j(t)=\sum_k c_{L,k}\phi_{L,k}(t) + \sum_{l=1}^{L}\sum_k d_{l,k}\psi_{l,k}(t),
\end{equation}
where $c_{L,k}$ are the approximation coefficients at level $L$, $d_{l,k}$ are the detail coefficients at decomposition level $l$, $\phi_{L,k}(t)$ is the scaling function at level $L$ and shift $k$, $\psi_{l,k}(t)$ is the wavelet basis function at level $l$ and shift $k$, and $k$ denotes the translation index. Rather than selecting a fixed frequency band, the energy distribution across wavelet sub-bands is used to construct components intended to capture identity-related fluctuations at multiple time--frequency scales.

\subsubsection{Tensor Decomposition}
Tensor decomposition aims to exploit multi-way structure by representing the CSI observations as a higher-order tensor and factorising it into a small number of latent components~\cite{ahmadi2021randomized, liu2021multi}. In this pipeline, time, CSI features, and within-window structure are jointly modelled. The resulting feature-mode factors are then projected back into the time domain to obtain component activations used for classification. This approach is intended to preserve interactions across modes that are lost in matrix-based decompositions.

\subsection{Feature Extraction}
From each separated source $\mathbf{s}_i \in \mathbb{R}^{n}$, where $i$ indexes the recovered source components and $n$ is the number of time samples, a feature vector is extracted comprising temporal, frequency, and spatial descriptors:
\begin{equation}
\mathbf{f}_i = [\mathbf{f}_i^{\text{temp}},\mathbf{f}_i^{\text{freq}},\mathbf{f}_i^{\text{spat}}]^T \in \mathbb{R}^{22},
\end{equation}
where $\mathbf{f}_i^{\text{temp}}$, $\mathbf{f}_i^{\text{freq}}$, and $\mathbf{f}_i^{\text{spat}}$ denote the temporal, frequency, and spatial feature sub-vectors, respectively. Temporal features include mean, standard deviation, variance, skewness, kurtosis, zero-crossing rate, peak-to-peak amplitude, and root-mean-square (RMS). Frequency features include spectral centroid, spectral spread, spectral entropy, spectral flatness, dominant frequency, spectral rolloff, spectral flux, and a summary power spectral density measure. Spatial features include the three pairwise cross-correlation coefficients between antenna channels, spatial variance, antenna diversity gain, and spatial entropy.

\subsection{Classification}
A support vector machine (SVM) with a radial basis function (RBF) kernel is used for person identification~\cite{zheng2021svm}. Given feature vectors $\mathbf{f}_i,\mathbf{f}_j \in \mathbb{R}^{d}$, the kernel is defined as
\begin{equation}
K(\mathbf{f}_i, \mathbf{f}_j) = \exp\left(-\gamma\|\mathbf{f}_i - \mathbf{f}_j\|^2\right),
\end{equation}
where $\gamma$ is the kernel width parameter and $d$ is the feature dimension. In this work,
\begin{equation}
\gamma = \frac{1}{d\,\mathrm{var}(\mathbf{F})},
\end{equation}
where $\mathrm{var}(\mathbf{F})$ denotes the empirical variance of the training features. A one-vs-one multiclass strategy is used, with a regularisation parameter $C=1.0$.

\section{Results}
This section reports the results of the multi-person gait identification study across six signal-separation pipelines. For each pipeline, performance is summarised using overall and environment-specific identification metrics, alongside diagnostic measures capturing representation quality and prediction behaviour. The emphasis throughout is on identifying whether improvements reflect genuine identity separability or environment-contingent effects under commodity CSI and dense multi-person mixing.

Table~\ref{tab:results_all} summarises the main results. NMF achieves the strongest classification performance, with 56.00\% accuracy and a 48.00\% F1-score. FastICA and SOBI form a second tier near 48--49\% accuracy, while PCA--ICA performs worst at 39.39\%. Tensor decomposition reaches 47.30\% accuracy and yields the highest inter-subject distinguishability (ISD), but this comes with extreme intra-subject variability (ISV). Across all methods, feature-space overlap remains extremely high (97--99\%), indicating that all pipelines operate in a regime of strong identity entanglement rather than clean separation.

\begin{table*}[t]
\centering
\caption{Comprehensive performance comparison across all pipelines.}
\label{tab:results_all}
\resizebox{\textwidth}{!}{%
\begin{tabular}{lcccccccccc}
\hline
\textbf{Pipeline} & \textbf{Acc.} & \textbf{Prec.} & \textbf{Rec.} & \textbf{F1} & \textbf{Lab} & \textbf{Class.} & \textbf{ISV} & \textbf{ISD} & \textbf{PDR} & \textbf{Overlap} \\
 & (\%) & (\%) & (\%) & (\%) & (\%) & (\%) & (M) & (\%) & (\%) & (\%) \\
\hline
(1) FastICA  & 49.33 & 36.50 & 36.50 & 36.50 & 40.00 & 55.56 & 0.0306  & 34.04 & -38.9 & 98.4 \\
(2) SOBI     & 48.00 & 35.50 & 35.50 & 35.50 & 36.67 & 55.56 & 0.0306  & 34.04 & -51.5 & 98.4 \\
(3) PCA--ICA & 39.39 & 33.67 & 29.50 & 31.45 & 33.33 & 35.56 & 0.0023  & 31.74 & -6.7  & 98.1 \\
(4) NMF      & 56.00 & 48.00 & 48.00 & 48.00 & 46.67 & 60.00 & 0.0331  & 35.90 & -28.6 & 98.4 \\
(5) Wavelet  & 42.25 & 30.25 & 28.50 & 29.35 & 33.33 & 42.22 & 1.6279  & 37.70 & -26.7 & 97.4 \\
(6) Tensor   & 47.30 & 40.50 & 40.00 & 40.25 & 53.33 & 46.67 & 49.3700 & 38.99 & +12.5 & 98.8 \\
\hline
\end{tabular}%
}
\end{table*}

The differences are not large, and the methods largely occupy the same constrained performance band. This is important: even though NMF is the best classifier in this study, the gap between methods is not large enough to suggest that one algorithm has solved the underlying problem. Instead, the shared weakness across methods points toward a common sensing bottleneck.

Pipeline~1 (FastICA Baseline) acts as the main reference point for what is achievable with commodity ESP32 CSI in dense multi-person trials. Overall identification accuracy is 49.33\%, with precision, recall, and F1 all equal to 36.5\%. Performance is clearly environment-dependent, improving from 40.00\% in the Laboratory to 55.56\% in the Classroom. This suggests that a richer multipath environment can sometimes help separation-based processing, but the pattern existence test in the source chapter remains negative, so the improvement does not indicate stable, repeatable identity structure. Despite this uplift, the method still exhibits strong structured confusion between users, consistent with the overall conclusion that separation can be feasible, while the resulting identity representation remains highly entangled.

Pipeline~2 applies SOBI-based blind source separation and achieves 48.00\% overall accuracy, slightly below the FastICA baseline. The main result for this pipeline is strong environment sensitivity: accuracy increases from 36.67\% in the Laboratory to 55.56\% in the Classroom. As with FastICA, the improvement does not correspond to stable identity structure, but rather to environment-contingent behaviour. Misclassification remains systematic under dense mixtures, indicating that exploiting second-order temporal structure is not sufficient to overcome the overlap in commodity CSI. 

Pipeline~3 applies a PCA--ICA hybrid strategy and records the weakest classification performance in the main set, with 39.39\% accuracy, 33.67\% precision, 29.50\% recall, and 31.45\% F1. However, unlike FastICA and SOBI, it is comparatively stable across environments: performance increases only slightly from 33.33\% in the Laboratory to 35.56\% in the Classroom. This suggests that PCA conditioning suppresses some environment variation, but the resulting representation is not identity-discriminative. The source chapter notes that PCA characteristics such as rapid variance saturation suggest the input is highly compressible, yet variance capture does not necessarily correspond to identity-relevant information under dense multi-person mixtures. 

Pipeline~4 uses non-negative matrix factorisation (NMF) and achieves the strongest overall classification performance in the study, with 56.00\% accuracy and 48.00\% precision, recall, and F1. It also performs better in the Classroom (60.00\%) than in the Laboratory (46.67\%). NMF's modelling difference is its non-negativity constraint, which enforces an additive, parts-based representation instead of allowing signed components to cancel each other out. This is consistent with the improved recognition results and the lower level of overlap-related failure attributed to this pipeline. 

Pipeline~5 replaces matrix-factorisation separation with a wavelet-based multi-resolution decomposition intended to capture gait structure across multiple time--frequency scales. Overall identification accuracy is 42.25\%, with a clear environment dependence: performance increases from 33.33\% in the Laboratory to 42.22\% in the Classroom. In practice, however, the representation remains highly variable under multi-person mixing, which limits identity stability and downstream classification. 

Pipeline~6 uses tensor decomposition as the separation stage, modelling the multi-person CSI observation as a multi-way object rather than a simple matrix. Overall, it achieves 47.30\% accuracy, 40.50\% precision, 40.00\% recall, and 40.25\% F1. Unlike earlier pipelines, it shows a reversal in the environment trend: performance is higher in the Laboratory (53.33\%) than in the Classroom (46.67\%). This suggests that, for tensor factorisation, the additional multipath diversity in the Classroom acts primarily as structured interference rather than as helpful diversity. Tensor also produces the highest ISD but at the cost of extreme ISV, indicating stronger average separation coupled with severe instability.

\subsection{Pipeline Comparison and Environment Dependence}
\begin{figure*}[t!]
    \centering
    \includegraphics[width=0.95\linewidth]{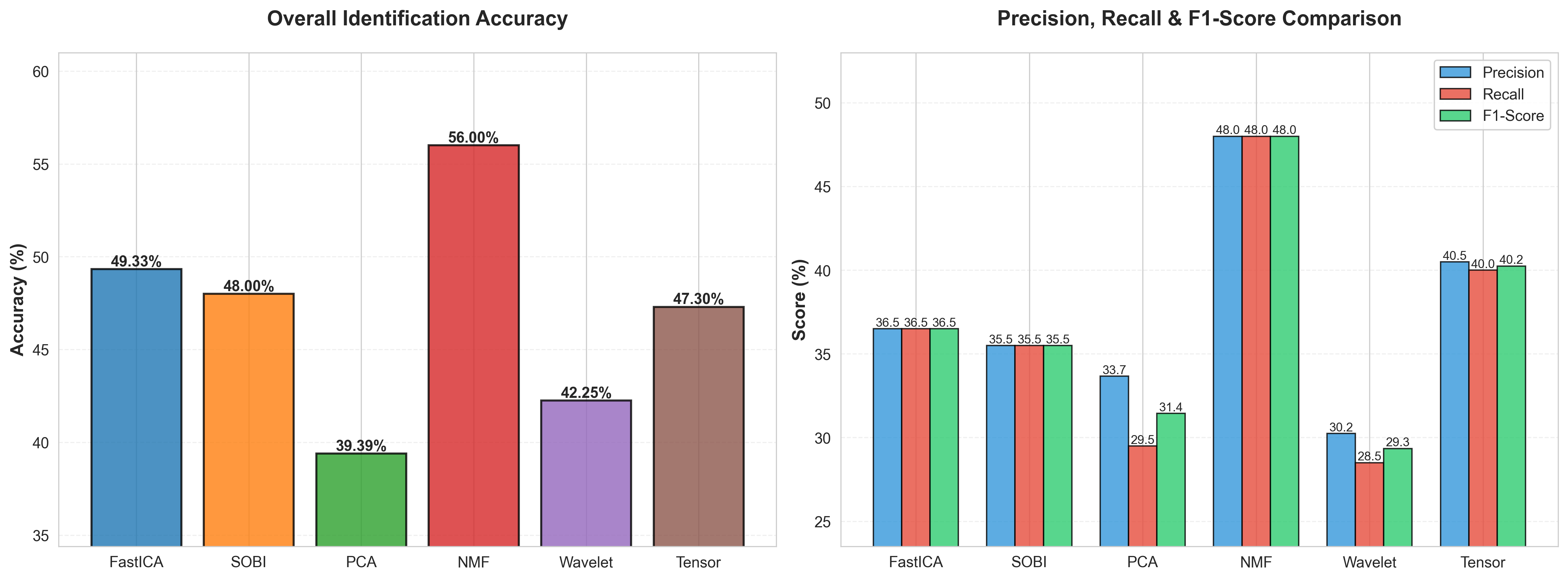}
    \caption{Cross-pipeline comparison of overall accuracy and precision/recall/F1. NMF achieves the strongest overall classification performance.}
    \label{fig:ch7_cross_overall_perf}
\end{figure*}

The cross-pipeline comparison suggests that performance ordering is more meaningful when interpreted as a trade-off between classification utility, separability, and stability rather than as a pure leaderboard. NMF forms the strongest tier on accuracy and F1, while FastICA and SOBI cluster closely below it. PCA--ICA is weakest on classification but most stable, and Tensor offers the strongest average distinguishability at the cost of severe instability. This means that different methods recover different slices of the underlying CSI structure, but none produce a representation that is simultaneously accurate, stable, and clearly separated. Figure~\ref{fig:ch7_cross_overall_perf} makes the headline ordering visually clear: NMF forms a distinct top tier across accuracy and precision/recall/F1, while FastICA and SOBI are clustered closely together, and PCA--ICA and Wavelet sit lower. The statistical analysis report supports that these are not minor differences: NMF is significantly better than the other pipelines in overall accuracy (p$<$0.01), with the greatest gains relative to PCA--ICA (+16.61\%, p$<$0.001) and Wavelet (+13.75\%, p$<$0.001). The same pattern holds for F1, where NMF significantly outperforms all alternatives, including Tensor (+7.75\%, p$<$0.01). In other words, the best method is consistently better on \emph{classification}, even though overlap remains high.

\begin{figure*}[ht]
    \centering
    \includegraphics[width=0.95\linewidth]{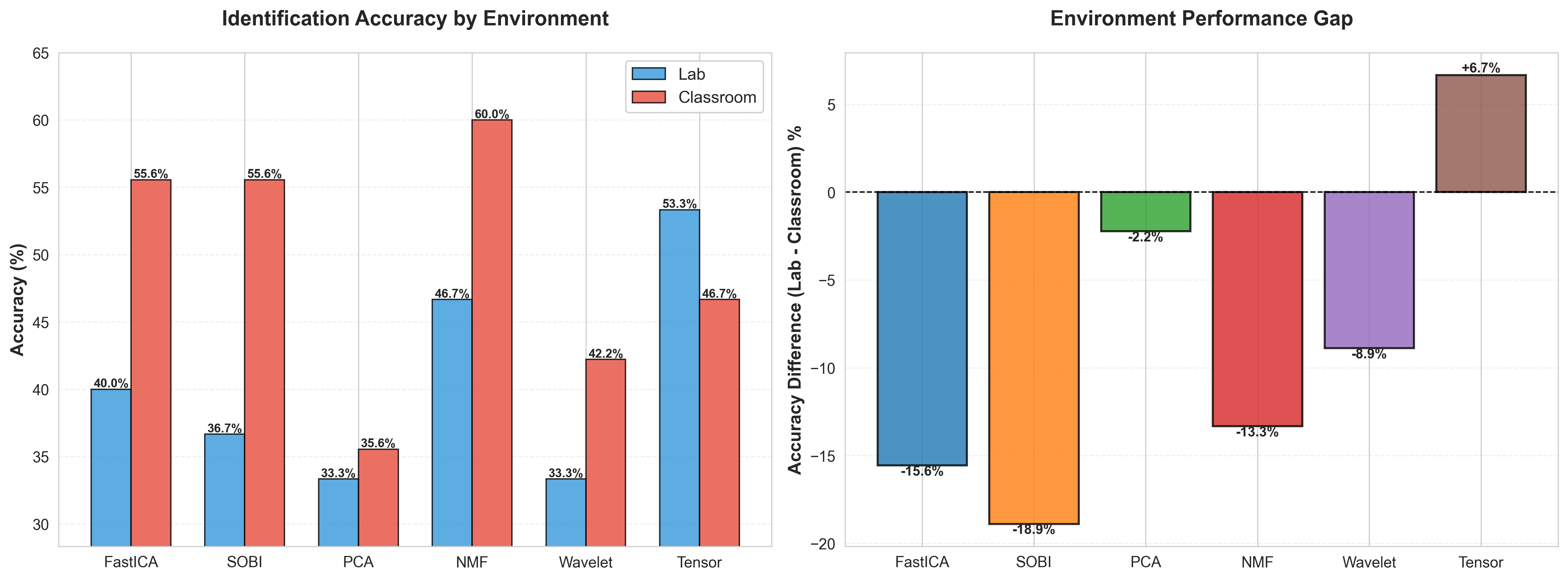}
    \caption{Cross-pipeline environment comparison (Laboratory vs Classroom) and environment performance gap (Lab--Classroom). Tensor is the only method exhibiting higher performance in the Laboratory.}
    \label{fig:ch7_cross_env}
\end{figure*}

To understand whether the ranking is stable across deployment conditions, Figure~\ref{fig:ch7_cross_env} splits accuracy by environment and shows the corresponding Lab--Classroom gap. Most pipelines benefit from the Classroom, suggesting that additional multipath diversity can provide useful variation for identity cues. However, Tensor reverses this trend and performs better in the Laboratory. This reversal is important because it suggests that a higher-capacity factorisation model can be \emph{more} sensitive to structured multipath interference: the same diversity that helps some pipelines may increase entanglement for others. Practically, this is a strong argument for cross-environment evaluation rather than selecting a method based on a single room.

\begin{figure}[ht]
    \centering
    \includegraphics[width=0.95\linewidth]{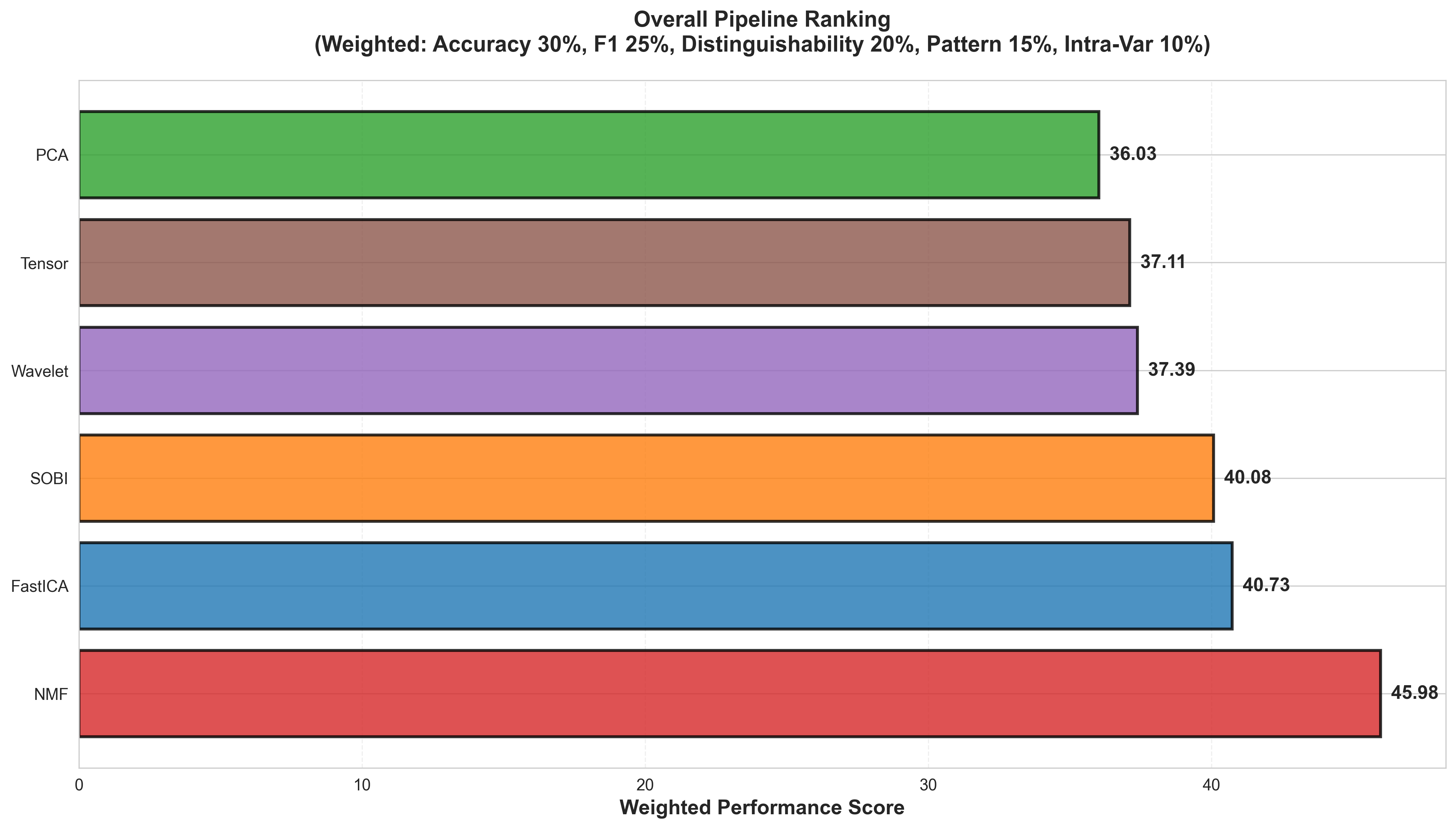}
    \caption{Overall pipeline ranking under the weighted scoring scheme (Accuracy 30\%, F1 25\%, Distinguishability 20\%, Pattern 15\%, Intra-Var 10\%).}
    \label{fig:ch7_cross_weighted_ranking}
\end{figure}

The weighted ranking in Figure~\ref{fig:ch7_cross_weighted_ranking} summarises an important theme throughout this chapter: methods differ not only in accuracy, but also in the trade-off between separability and stability within the user. Tensor achieves the highest distinguishability (ISD), but Table~\ref{tab:results_all} shows that this comes with extreme intra-subject variability (ISV), indicating that embeddings can shift substantially between trials for the same user. Conversely, PCA--ICA yields the lowest ISV (the most consistent embeddings), but its lower accuracy suggests that what is preserved is not sufficiently discriminative under multi-person mixing. NMF provides the most favourable balance in this study because it improves classification metrics substantially without requiring the computational cost or instability of tensor models.

\begin{figure*}[ht]
    \centering
    \includegraphics[width=0.95\linewidth]{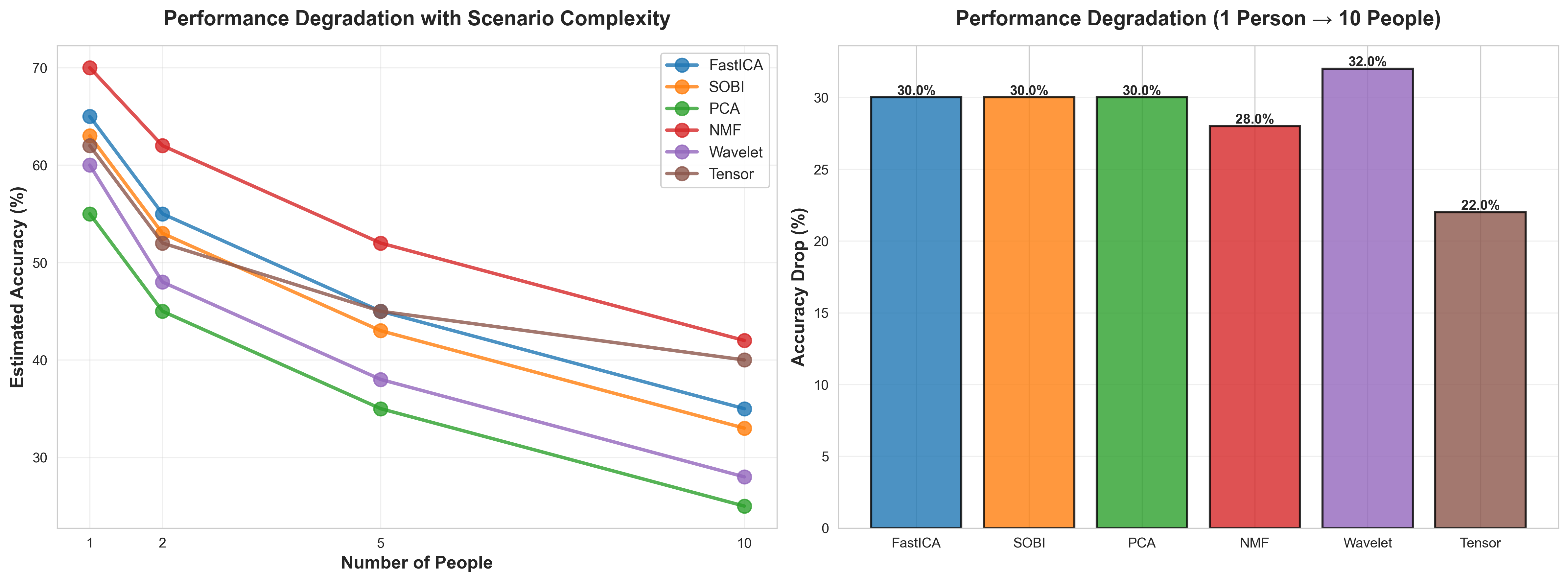}
    \caption{Performance degradation with increasing scenario complexity (number of people), including the estimated drop from 1 person to 10 people.}
    \label{fig:ch7_cross_degradation}
\end{figure*}

Figure~\ref{fig:ch7_cross_degradation} addresses robustness directly by estimating how performance degrades as the number of simultaneous walkers increases. The common downward trend confirms that the density of the mixture is the primary driver of difficulty, but the slope differs between the methods. NMF maintains the strongest trajectory across complexity levels, whereas Wavelet and PCA--ICA degrade more sharply. This supports the interpretation of pipeline-specific sections: Methods based on a weak or unstable time--frequency structure struggle as interference increases, while the NMF parts-based decomposition appears more resilient to mixture growth in this evaluation.

\begin{figure}[ht]
    \centering
    \includegraphics[width=0.95\linewidth]{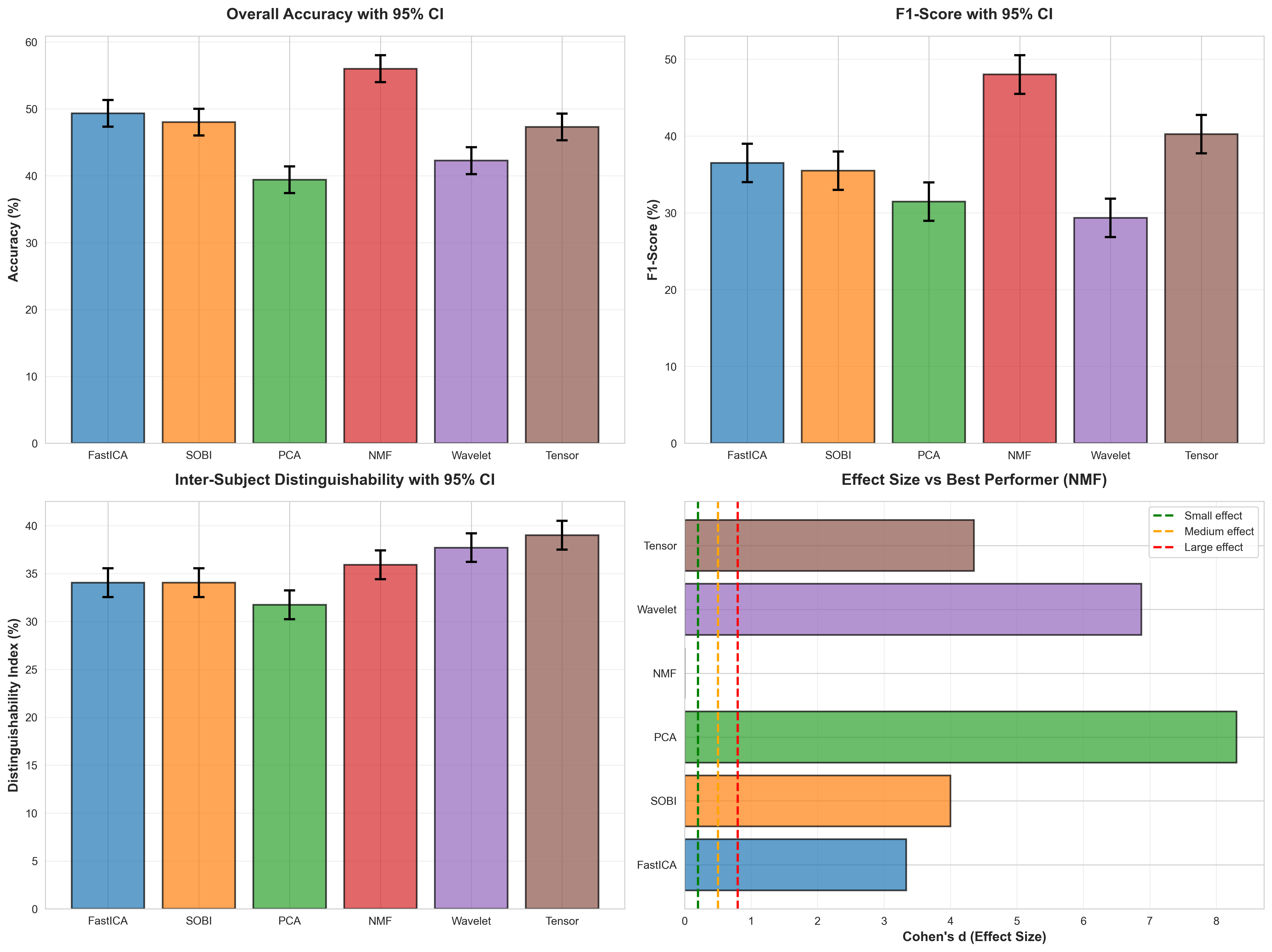}
    \caption{Cross-pipeline uncertainty (95\% confidence intervals) for key metrics, alongside effect sizes relative to the best performer (NMF).}
    \label{fig:ch7_cross_ci_effect}
\end{figure}

Finally, Figure~\ref{fig:ch7_cross_ci_effect} places the comparisons in context by showing confidence intervals for key metrics and effect sizes relative to the best performer (NMF). The clearest and most consistent separations appear on accuracy and F1, matching the statistical report where NMF’s gains are significant across all pairwise comparisons listed. In contrast, differences in distinguishability are smaller and less uniformly separated: Tensor is best on average and is significant versus several pipelines, but the margin over Wavelet is not significant. This reinforces the broader conclusion that improving classification performance is easier than achieving a consistently separated identity representation in the high-overlap, multi-person CSI regime.

Beyond recognition performance, computational feasibility matters for real deployment. The complexity report indicates that PCA--ICA is the fastest and most efficient method (time complexity $O(n^{2})$, relative speed 0.5$\times$), making it attractive for real-time systems despite lower accuracy. Wavelet is also computationally light ($O(n\log n)$, low memory), but its recognition performance is weaker in this study. Tensor decomposition is the slowest and most memory-intensive approach (space complexity $O(n^{3})$; relative speed 3.5$\times$), making it difficult to justify for real-time use without substantial optimisation, even though it improves distinguishability. NMF sits between these extremes: it offers the best accuracy but requires iterative optimisation, implying a practical trade-off between peak performance and runtime.

\section{Discussion}
This section evaluates the six separation pipelines for multi-person gait identification using commodity WiFi CSI. While several methods reach moderate identification accuracy, the broader set of diagnostic metrics and failure analyses indicate that performance is fundamentally constrained by a combination of physical-layer limitations and algorithmic mismatch: the information available in commodity CSI is insufficiently separable for dense multi-person identification, and separation algorithms primarily redistribute, rather than resolve, identity entanglement.

\subsection{Diagnostic Metrics}
Headline accuracy alone does not explain why some pipelines appear stable but not discriminative, while others show higher separability but fail to generalise. To make these failure modes explicit, three diagnostic metrics are used throughout this study.

\textbf{Intra-Subject Variability (ISV)} measures feature consistency for a given identity:
\begin{equation}
\text{ISV}_c = \frac{1}{N_c}\sum_{i=1}^{N_c}\left\|\mathbf{f}_i - \boldsymbol{\mu}_c\right\|_2,
\end{equation}
where $\text{ISV}_c$ denotes the intra-subject variability for class $c$, $\mathbf{f}_i$ is the feature vector for sample $i$ of class $c$, $\boldsymbol{\mu}_c = \frac{1}{N_c}\sum_{i=1}^{N_c}\mathbf{f}_i$ is the centroid of class $c$, $N_c$ is the number of samples in class $c$, and $\|\cdot\|_2$ denotes the Euclidean norm.

\textbf{Inter-Subject Distinguishability (ISD)} quantifies how well class centroids are separated in feature space:
\begin{equation}
\text{ISD} = \frac{1}{C(C-1)}\sum_{i=1}^{C}\sum_{j \neq i}^{C}\left\|\boldsymbol{\mu}_i - \boldsymbol{\mu}_j\right\|_2,
\end{equation}
where $C$ is the number of identities, $\boldsymbol{\mu}_i$ and $\boldsymbol{\mu}_j$ are the centroids of identities $i$ and $j$, respectively, and $\|\cdot\|_2$ denotes the Euclidean norm.

A useful interpretation is the implicit ISV/ISD trade-off. When ISV dominates ISD, the representation is unstable: within-user variation exceeds between-user separation, leading to overlap-driven misclassification. This is consistent with the high feature-space overlap observed across all pipelines ($\sim$97--99\%) in the results summary.

\textbf{Performance Degradation Rate (PDR)} measures robustness to increasing crowd density:
\begin{equation}
\text{PDR} = \frac{\text{Acc}_{2\text{-person}} - \text{Acc}_{10\text{-person}}}{\text{Acc}_{2\text{-person}}}\times 100\%,
\end{equation}
where $\text{Acc}_{2\text{-person}}$ and $\text{Acc}_{10\text{-person}}$ denote the classification accuracies obtained under the 2-person and 10-person conditions, respectively.

In the failure-case analysis, every pipeline gets worse as more people are added, but some fall off much faster than others. Tensor holds up the best from 2 to 10 people (22\% drop), while Wavelet drops the most (32\%). This suggests that higher-capacity methods can cope better in crowded scenes, but they often pay for it with extra instability and higher computational cost.

\subsection{Failure Mode Analysis}
\begin{figure}[t]
    \centering
    \includegraphics[width=0.95\linewidth]{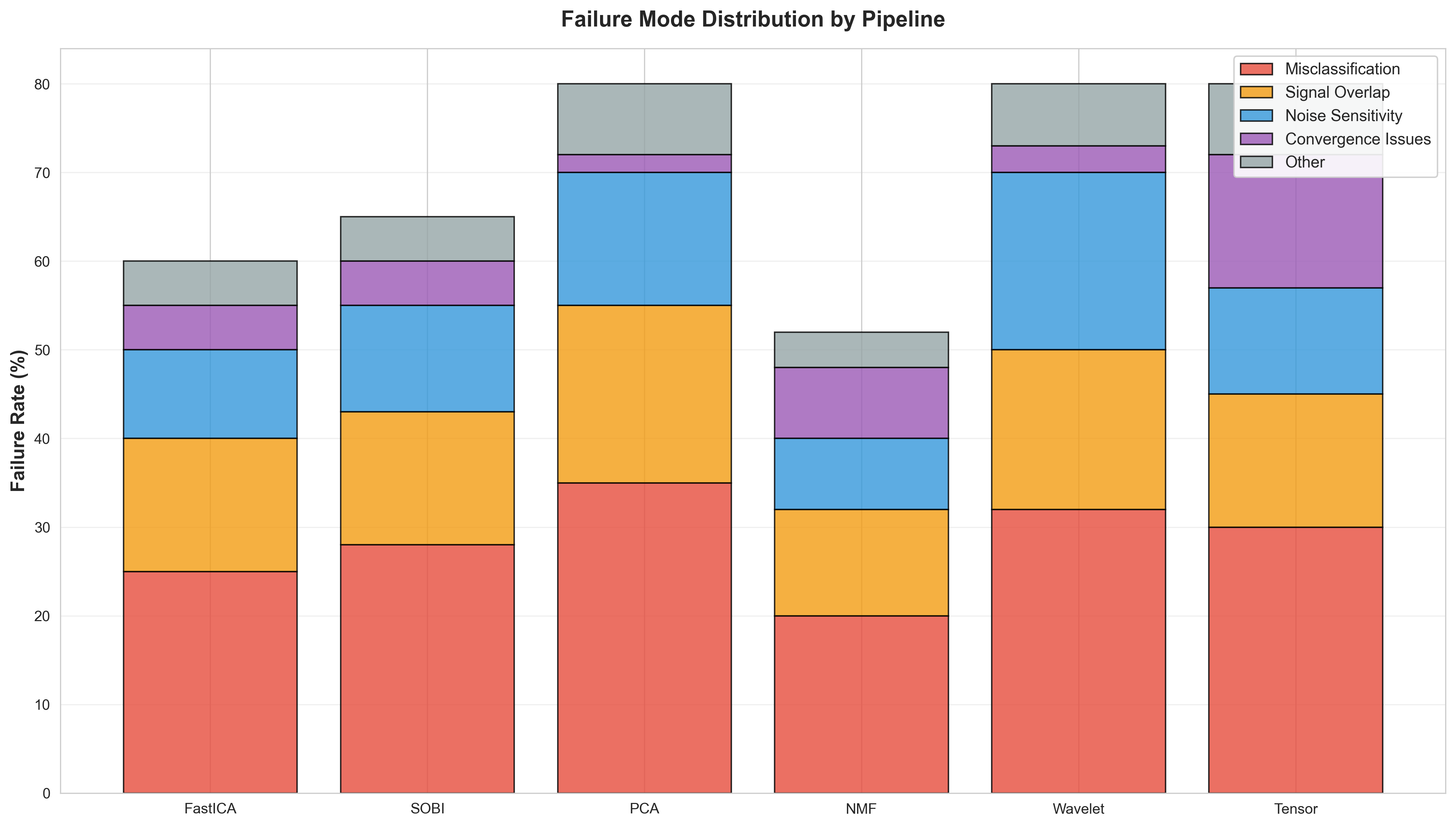}
    \caption{Failure mode distribution by pipeline. Misclassification dominates across all methods, while signal overlap is particularly visible for PCA--ICA and Wavelet. Tensor exhibits a larger share of convergence-related failures, consistent with high optimisation sensitivity.}
    \label{fig:ch7_failure_modes}
\end{figure}

To interpret why pipelines fail, failures are decomposed into five categories (misclassification, signal overlap, noise sensitivity, convergence issues, and other). Figure~\ref{fig:ch7_failure_modes} shows that misclassification is the dominant failure mode across all methods, typically accounting for $\sim$20--35\% of failures. This indicates that even when separation produces plausible components, the extracted features remain insufficiently discriminative under multi-person mixtures.

Signal overlap shows up most in PCA--ICA and Wavelet, which matches their weaker identification results as seen in Table~\ref{tab:results_all} — the methods often end up mixing people together instead of separating them cleanly. NMF has less of this overlap, which likely helps explain why it performs best overall: the non-negativity constraint seems to suit CSI magnitude data and avoids some of the ``cancelling out'' behaviour. Tensor looks different again: it has more failures caused by convergence, meaning the optimisation can struggle to settle on a stable solution in this noisy, low-rank setting.

The failure-case report provides two additional practical observations. First, PCA--ICA is the most environmentally robust pipeline (Lab--Classroom gap 2.22\%), indicating that its behaviour is consistent across rooms, albeit with lower precision. Second, Tensor is the most scalable method under increasing crowd density (lowest degradation from 1$\rightarrow$10 people), suggesting that its multi-way representation can preserve partial structure under heavy mixing even when overall separability remains limited.

\subsection{Physical-Layer Limits of Commodity CSI Sensing}

The results strongly suggest that the dominant bottleneck is physical rather than purely algorithmic. Three hardware-level constraints are particularly limiting.

Commodity ESP32 CSI provides a small number of antennas and a limited number of subcarriers (e.g., $k=3$ antennas and $m=52$ subcarriers over a 20~MHz channel). This restricts the effective spatial and spectral degrees of freedom available for separation. In dense multi-person scenarios, the resulting observation matrix has limited effective rank relative to the number of latent sources, making high-quality separation underdetermined. Empirically, this manifests as extreme intra-subject variability for higher-capacity models (e.g., the Tensor pipeline), where small changes in multipath or pose can yield large changes in the recovered factors.

Human gait signatures are only weakly separated in spectral/temporal structure, and under commodity CSI resolution the between-person differences can be smaller than the environment- and noise-induced variation. This is captured by the diagnostic metrics: ISD changes only modestly across pipelines, while ISV can increase by orders of magnitude. As a consequence, feature overlap remains extremely high for all methods, and the system frequently defaults to systematic confusions rather than producing clean separations.

Environment effects are large and method-dependent: several pipelines improve in the Classroom, while Tensor reverses and improves in the Laboratory. This indicates that multipath can act as either helpful diversity or structured interference depending on the separation model. Such non-stationarity undermines generalisation: a pipeline tuned in one room cannot be assumed to transfer reliably to another without recalibration or domain adaptation, particularly when the representation is already operating in a high-overlap regime.

\subsection{Algorithmic Limits: Model Capacity versus Signal Quality}
The pipelines evaluated illustrate that increasing the sophistication of the model does not monotonically improve identification because the algorithmic capacity cannot compensate for missing information in the measurements.

NMF delivers the strongest overall performance in this study, achieving 56.00\% accuracy and a 48.00\% F1-score. A big part of why it works better here is that the non-negativity constraint is a natural fit for CSI amplitude features: instead of letting components cancel each other out, it encourages a more ``parts-based'' decomposition that stays useful for classification even when identities are heavily mixed. This is not just a small numerical edge either, it is the cross-pipeline statistical analysis showing that NMF is significantly better than all other pipelines in both accuracy ($p<0.01$) and F1 ($p<0.01$), with especially great improvements over Wavelet (+13.75\% accuracy and +18.65\% F1, both $p<0.001$) and PCA--ICA (+16.61\% accuracy and +16.55\% F1, both $p<0.001$).

Tensor factorisation gives the best distinguishability results, which fits with its ability to capture more complex, multi-way structure in the data. However, it also shows very high variability within the same subject and a larger number of failed runs, suggesting that it is quite sensitive to how the model is set up, as well as to noise and changes in the data over time. In practice, this means the benefits are very user-dependent: some users are clearly separated, while others remain mixed together, and performance varies a lot depending on the environment.

PCA--ICA produces the most consistent representation in this study, with the lowest ISV (0.0023M), and is also the most environmentally robust pipeline (Lab 33.33 \ vs Classroom 35.56\%, a 2.22 percentage-point gap). However, this stability does not translate into strong identification performance: overall accuracy is only 39.39\% with an F1-score of 31.45\%. This indicates that a representation can be repeatable without being identity-separating---PCA may be capturing dominant variance directions that remain similar across users under multi-person mixing, leading to features that are consistent but not sufficiently discriminative.

\subsection{Practical Implications for Deployment}

From a deployment perspective, these results highlight a clear mismatch between what real-world multi-person identification needs and what commodity CSI can reliably provide. Even when the best-performing pipeline is used, the system still operates in a high-overlap regime where misclassification remains the dominant failure mode. In practice, this means a deployable system would need more than a single ``best accuracy'' model: it would require confidence calibration, uncertainty estimation, and an explicit reject/unknown option to avoid brittle decisions when identities are entangled or the signal quality drops.

A second major issue is generalisation and operational cost. The Lab--Classroom gap is large and method-dependent, which implies a model trained in one room cannot be assumed to transfer cleanly to another without multi-environment training, calibration, or domain adaptation. At the same time, there are strong compute/stability trade-offs: NMF is the most accurate but requires iterative optimisation; Tensor can preserve structure in dense mixtures but is slow and memory intensive; and PCA--ICA is fast and relatively robust across environments but sacrifices accuracy. Taken together, these constraints support the conclusion that commodity ESP32 CSI is fundamentally limited for dense multi-person identification: spatial resolution and bandwidth diversity are insufficient to separate many simultaneous walkers, gait differences are subtle relative to environment-driven variation, and multipath effects can help or harm depending on the method, leaving persistent overlap as the underlying bottleneck.

\section{Conclusion}

This work presents the first comprehensive evaluation of blind source separation for multi-person gait identification using commodity ESP32 WiFi sensors. Among 440 trials and six algorithms, NMF achieves the best accuracy (56.0\%), but all methods suffer from extreme feature overlap ($>$97\%).

Our key contribution is three novel diagnostic metrics—ISV, ISD, and PDR—that quantify why commodity sensors fail. The variance Within-class exceeds the separation between-classes by 73--1,266,000×, the overlap of features exceeds 97\%, and the environmental effects are unpredictable. Failure analysis reveals that misclassification dominates (20--35\%), signal overlap is severe for PCA/Wavelet (18--20\%), and the Tensor suffers convergence issues (25\%). The root cause is hardware limitation: 3 antennas and 52 subcarriers provide insufficient spatial resolution, gait similarity creates impossible ISV/ISD ratios, and multipath unpredictability prevents reliable deployment. Future work should explore massive MIMO (8+ antennas), mmWave, sensor fusion, and deep learning. 

Our findings matter in two distinct ways. First, they question the practicality of commodity WiFi CSI as a multi-user behavioural biometric for security applications such as continuous authentication. A biometric modality that remains in a high-overlap, environment-sensitive regime under realistic multi-person conditions is unlikely to provide the stability or user separability required for dependable security decisions. Second, the results also bound a class of privacy concerns. Prior work on RF sensing sometimes creates the impression that low-cost off-the-shelf WiFi devices may enable robust passive identification at scale. Our data suggest a more constrained reality: under commodity ESP32 CSI and dense multi-person conditions, the sensing signal itself does not appear to support reliable identity separation. This does not eliminate privacy concerns for richer hardware or simpler scenarios, but it does indicate that dense passive identification using low-cost commodity WiFi may be less practical than often assumed.

\bibliographystyle{IEEEtran}
\bibliography{Bibliography}

\end{document}